# Air Quality PM2.5 Index Prediction Model Based on CNN - LSTM

Zicheng Guo [1,3], Shuqi Wu [2,4], Meixing Zhu[1,5], He Guandi*[1,6]
[1] Guizhou University, Guizhou, China
[2] Amazon, New York, USA

[3] agr.zcguo23@gzu.edu.cn

[4] shuqiwu2023@u.northwestern.edu

[5] 1271439328@qq.com

[6] gdhe@gzu.edu.cn

**Abstracts.** With the intensification of global climate change, accurate prediction of air quality indicators, especially PM2.5 concentration, has become increasingly important in fields such as environmental protection, public health, and urban management. To address this, we propose an air quality PM2.5 index prediction model based on a hybrid CNN-LSTM architecture. The model effectively combines Convolutional Neural Networks (CNN) for local spatial feature extraction and Long Short-Term Memory (LSTM) networks for modeling temporal dependencies in time series data. Using a multivariate dataset collected from an industrial area in Beijing between 2010 and 2015—which includes hourly records of PM2.5 concentration, temperature, dew point, pressure, wind direction, wind speed, and precipitation—the model predicts the average PM2.5 concentration over 6-hour intervals. Experimental results show that the model achieves a root mean square error (RMSE) of 5.236, outperforming traditional time series models in both accuracy and generalization. This demonstrates its strong potential in real-world applications such as air pollution early warning systems. However, due to the complexity of multivariate inputs, the model demands high computational resources, and its ability to handle diverse atmospheric factors still requires optimization. Future work will focus on enhancing scalability and expanding support for more complex multivariate weather prediction tasks.

## 1. Introduction

With the ongoing changes in the global climate, extreme weather events and frequent meteorological disasters have posed serious threats to societal operations and the ecological environment. Accurate meteorological forecasting plays a vital role in various fields such as disaster prevention and mitigation, agricultural production, and urban transportation. Among traditional forecasting approaches, statistical regression-based methods have long been applied to weather prediction tasks due to their simplicity and interpretability. These models—such as multiple linear regression and support vector regression (SVR)—learn relationships between meteorological variables and target values from historical data. However, they often struggle with the nonlinear and highly dynamic nature of weather systems, resulting in limited prediction accuracy and poor generalization capabilities.

In recent years, the rapid development of deep learning has introduced new opportunities in atmospheric prediction. Deep neural networks are particularly adept at capturing complex nonlinear dependencies and spatiotemporal patterns. This study proposes a CNN-LSTM-based air quality PM2.5 index prediction model, designed to enhance prediction accuracy of pollution concentration by combining the strengths of two neural architectures. The convolutional neural network (CNN) component is effective at extracting local spatial features from multi-dimensional weather data, while the long short-term memory (LSTM) network captures temporal dependencies across sequences, overcoming issues such as vanishing gradients and short-term memory in traditional RNNs.

To validate the model, we utilize a comprehensive air quality dataset from an industrial region in Beijing, covering hourly weather and pollution data from 2010 to 2015. This includes timestamped PM2.5 levels, temperature, dew point, pressure, wind direction, wind speed, and cumulative rain/snow hours. The prediction task focuses on forecasting the average PM2.5 concentration every 6 hours. Experimental results demonstrate that our model achieves a root mean squared error (RMSE) of 27.416 and a mean squared error (MSE) of 5.236, outperforming traditional regression-based methods in terms of both accuracy and robustness. Nonetheless, challenges remain in handling high-dimensional and multi-source data efficiently, which will be addressed in future work.

## 2. Related work

In recent years, with the rapid development of deep learning technology, prediction models based on deep neural networks have been continuously proposed and have achieved significant improvements in multiple fields. Especially in atmospheric pollution prediction tasks, deep learning methods, with their powerful feature extraction and temporal modeling capabilities, can more effectively capture the complex nonlinear relationship between weather and pollution indicators, improving the shortcomings of traditional methods in modeling ability and response speed.

Fei Xiao et al [1]. proposed an improved deep learning model (WLSTME) for daily PM2.5 concentration prediction. It first uses MLP to generate weighted historical PM2.5 data from neighbor sites based on distance, pollution concentration and wind conditions. Then, LSTM is applied to extract spatiotemporal features. Finally, another MLP integrates these features with meteorological data. Experiments show WLMEST outperforms existing methods.

Jian Peng et al [2]. conducted a study on PM2.5 concentration prediction. They collected one - year data of six main meteorological parameters and PM2.5 concentrations at two sites in central China's Hunan Province. These data trained, validated, and evaluated XGBoost and fully connected neural network models. After parameter tuning and performance comparison, XGBoost showed better prediction ability, especially in the nighttime test dataset. The study also analyzed how meteorological variables affect PM2.5 concentrations.

In their study, Inchoon Yeo et al [3]. proposed an integrated deep learning model combining CNN and GRU to forecast PM2.5 concentrations at 25 Seoul stations. Trained on 2015–2017 data from multiple sources, the model predicted 2018 concentrations. They also introduced a geographical polygon group model to optimize neighboring stations for better accuracy. Compared to single - station data methods, this approach improved PM2.5 prediction accuracy by about 10% on average, offering an efficient deep - learning - based solution.

## 3.Data Introduction

The dataset used in this study comes from an air pollution industrial zone in Beijing, covering 2010–2015 with hourly records of atmospheric indicators, including PM2.5 concentration, temperature, pressure, wind data, and precipitation. To ensure data quality, missing or invalid values (-99/NaN) were removed, and data was reorganized into 6-hour time windows for average PM2.5 forecasting. All variables were normalized, and new features were derived through time correlation and coupling analysis. This preprocessing pipeline enhanced data expressiveness and consistency, providing robust input for the CNN-LSTM model to achieve accurate PM2.5 predictions.

**Table 1.** Variables and descriptions

| variable | description |
| --- | --- |
| year | year of data in this row |
| month | month of data in this row |
| day | day of data in this row |
| hour | hour of data in this row |
| PM2.5 | PM2.5 concentration |
| DEWP | Dew Point |
| Temperature | Temperature |
| PRES | Pressure |
| cbwd | Combined wind direction |
| weather | Meteorological conditions |
| Iws | Cumulated wind speed |

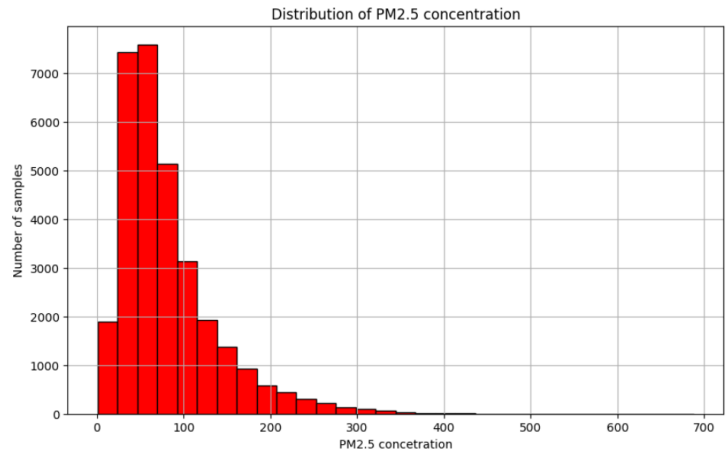

**Figure 1.** Frequency Distribution of PM2.5 Concentration Levels

Figure 1 presents the distribution of PM2.5 concentrations within the dataset used for air quality prediction. The histogram reveals a right-skewed distribution, with the majority of PM2.5 concentration values clustered at lower levels. Specifically, a significant peak is observed between 0 and 50μg/m³, indicating frequent occurrences of moderate pollution levels. As the concentration increases beyond 100μg/m³, the frequency of samples decreases sharply, reflecting less common severe pollution events. A long tail extends to the right, with very few samples exceeding 400μg/m³. This distribution pattern highlights the typical air quality conditions in the region, with occasional spikes in pollution. Such information is crucial for understanding the baseline air quality and for developing models that can effectively predict both regular and extreme pollution events.

## *4.*CNN-LSTM model

*4.1 CNN*

In the CNN-LSTM hybrid model proposed in this study, the Convolutional Neural Network (CNN) serves as the feature extraction module and plays a critical role in the task of PM2.5 concentration prediction. CNN is a type of feed-forward neural network that excels at processing data with a grid-like structure, such as images or multivariate time series data. Its core advantages—local receptive fields and weight sharing—enable efficient and accurate handling of complex environmental data with spatial and temporal correlations [4].

This study utilizes a multivariate air quality dataset collected from an industrial area in Beijing from 2010 to 2015. The dataset contains hourly records of PM2.5 concentration, temperature, dew point, air pressure, wind speed, wind direction, and cumulative precipitation and snowfall duration. Given the strong spatiotemporal dependencies inherent in this high-dimensional data, CNN is employed to automatically extract local spatial features from the raw multivariate time series. Specifically, the convolutional layers apply sliding convolution operations with various kernel sizes to detect representative local patterns, such as abrupt shifts in PM2.5 concentration under specific meteorological conditions or local interactions among weather factors.

For instance, under high humidity and low wind speed, PM2.5 levels tend to spike significantly. CNN can effectively identify such joint feature patterns via its convolutional filters, providing high-level representations as input for subsequent LSTM-based temporal modeling. Furthermore, pooling layers following the convolutional layers downsample the feature maps to reduce dimensionality and computational complexity while retaining critical features, thereby improving the model's training efficiency and generalization performance.

In summary, the integration of CNN in the proposed model significantly enhances the ability to model local features within multivariate air quality data. It also provides more discriminative and high-quality feature inputs for the LSTM module, which ultimately improves the accuracy and robustness of PM2.5 concentration prediction.

*4.2 LSTM*

As a specialized type of recurrent neural network, the Long Short-Term Memory (LSTM) network excels in handling time series data. Its core advantage lies in its ability to capture long-range temporal dependencies while effectively avoiding the vanishing gradient problem commonly encountered in traditional RNNs during training. In the CNN-LSTM-based air quality PM2.5 index prediction model proposed in this study, the LSTM plays a critical role in modeling the dynamic temporal trends within the time series, making it particularly suitable for prediction scenarios where there are significant temporal correlations between complex meteorological factors and pollutant concentrations [5].

LSTM, on the other hand, is primarily used to model long-term dependencies in time-series data by introducing three key gating structures; the forgetting gate determines the amount of information that needs to be discarded based on the current input and previous memory states. The input gate then determines what new information is worth being written to the state. Finally, the output gate combines the updated state and the current input to generate a prediction, which to some extent solves traditional gradient vanishing and gradient explosion problems when dealing with long sequential data RNNs. In this study, two layers of LSTM are by setting stacked the parameter values of units to LSTM 64 and 60 and return_sequences=True in the two models respectively. In addition, are also two fully connected layers used, the first containing 30 neurons and the second containing 10 neurons, and the activation functions are both ReLU, so that the last fully connected layer outputs only 1 value, which represents the predicted temperature and the Lambda layer scales the output value to the original temperature range. The period training epoch is set to 50.

*4.3 Model Training*

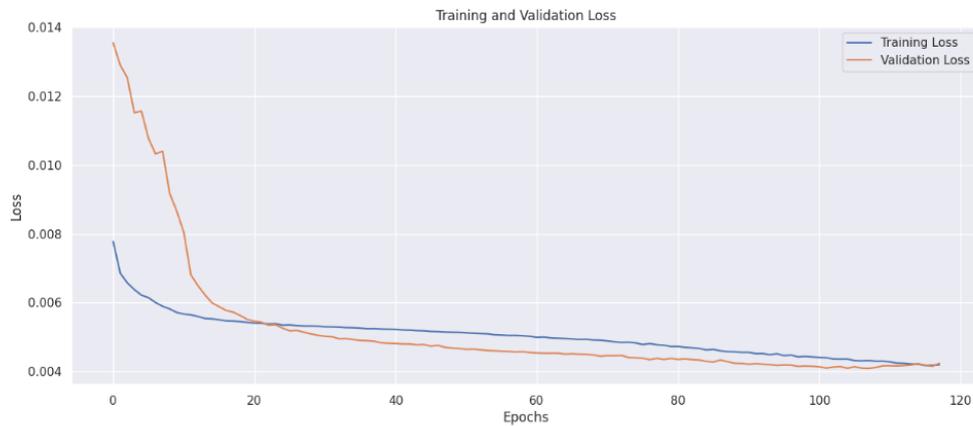

**Figure 2.** Training loss

Figure 2 presents the Training and Validation Loss curves for the CNN-LSTM model used in the PM2.5 concentration forecasting task. The x-axis denotes Epochs, and the y-axis denotes Loss. Initially, at Epoch 0, the Training Loss starts at approximately 0.014 and the Validation Loss at around 0.013. As training progresses, both losses show a downward trend. By Epoch 20, the Training Loss drops to roughly 0.006 and the Validation Loss to about 0.007. After Epoch 60, both curves plateau near 0.004, indicating convergence. The close values of Training and Validation Loss suggest the model generalizes well without overfitting. These loss values and the trend demonstrate the CNN-LSTM model's effective learning and capability in accurately forecasting PM2.5 concentrations.

**5. Analysis of results**

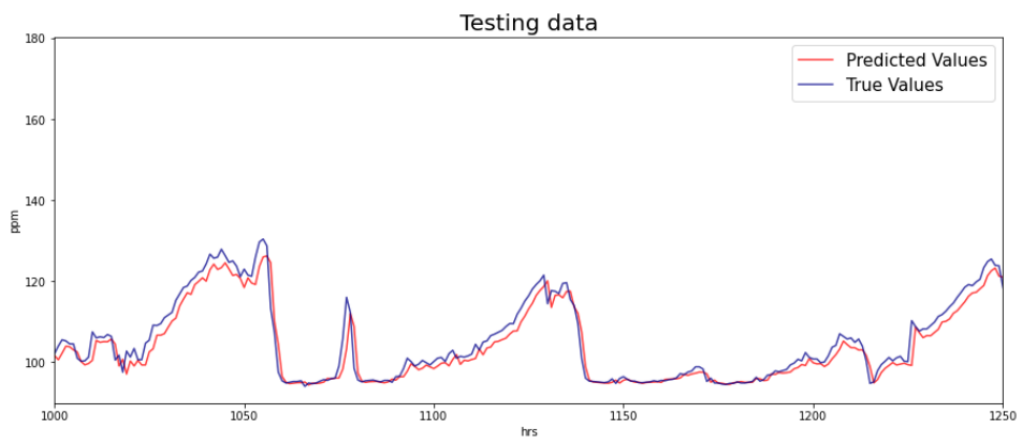

**Figure 3.** Ground Truth vs Forecast Comparison Over Time

Figure 3 illustrates a comparative analysis between the actual PM2.5 concentration levels (Ground Truth) and the predicted concentrations by the CNN-LSTM model during the testing phase. The visual representation indicates a strong alignment between the predicted and actual values, confirming the model's high predictive accuracy. Specifically, the CNN-LSTM architecture effectively captures the complex nonlinear patterns and temporal relationships present within the dataset, allowing for accurate tracking of general trends and fluctuations in pollution levels.

The x-axis represents time in hours, and the y-axis shows PM2.5 concentration in ppm. The red line indicates predicted values, while the blue line represents true values. Overall, the predicted values closely follow the trend of the true values across the entire time span, capturing both the general pattern and several significant peaks and troughs. However, some deviations are noticeable during specific periods—such as near hour 1050 and hour 1150—where the predicted values slightly lag or diverge from the true values. These discrepancies may be attributed to sudden fluctuations or complex environmental interactions not fully captured by the model. With an MSE [6] of 5.236 and RMSE [7] of 27.416, the model demonstrates reasonable accuracy and reliability. Despite minor deviations, the model provides a robust framework for PM2.5 forecasting, offering valuable support for air quality warning and environmental monitoring systems. Given the comprehensive data preprocessing, including the handling of missing values, normalization, and feature engineering based on time-series analysis, the model effectively leverages the high-quality input data to deliver meaningful predictions. Future improvements could focus on refining the model's response to abrupt environmental changes to enhance prediction precision further.

## 5. Summary

In this study, we proposed a CNN-LSTM-based hybrid deep learning model for PM2.5 concentration prediction, targeting improved forecasting accuracy and stability in multivariate, low-quality meteorological data scenarios. The model leverages the spatial feature extraction capabilities of CNN and the temporal dependency modeling strength of LSTM, enabling effective prediction of 6-hour moving average PM2.5 levels.

The dataset, collected from 2010 to 2015 in an industrial area of Beijing, consists of hourly records of PM2.5 concentration, dew point, temperature, pressure, wind direction, wind speed, and precipitation duration. A robust preprocessing pipeline was applied, including removal of invalid values (-99/NaN), time window restructuring, normalization, and the construction of derived features through correlation analysis. This improved data quality and model interpretability.

Experimental results demonstrate the effectiveness of the CNN-LSTM model, achieving a Mean Squared Error (MSE) of 5.236 and Root Mean Square Error (RMSE) of 27.416 on the test set, outperforming conventional time-series models. The model shows strong generalization under noisy environmental conditions, preserving key temporal trends and structural detail.

However, some limitations remain. The model requires substantial computational resources due to the high dimensionality of input features. Additionally, performance in extreme weather-induced PM2.5 fluctuations needs enhancement. Future work will aim to (1) develop lightweight model structures for real-time applications, (2) integrate external data sources such as remote sensing or geographic information for improved spatial awareness, and (3) extend the forecasting scope to other pollutants and time scales. These improvements will support the development of a more scalable and practical air quality early warning system.